\documentclass[letterpaper]{article} % DO NOT CHANGE THIS
\usepackage{aaai2026}  % DO NOT CHANGE THIS
\nocopyright
\usepackage{times}  % DO NOT CHANGE THIS
\usepackage{helvet}  % DO NOT CHANGE THIS
\usepackage{courier}  % DO NOT CHANGE THIS
\usepackage[hyphens]{url}  % DO NOT CHANGE THIS
\usepackage{graphicx} % DO NOT CHANGE THIS
\usepackage{natbib}  % DO NOT CHANGE THIS AND DO NOT ADD ANY OPTIONS TO IT
\usepackage{caption} % DO NOT CHANGE THIS AND DO NOT ADD ANY OPTIONS TO IT
\usepackage{algorithm}
\usepackage{algorithmic}

\usepackage{newfloat}
\usepackage{listings}
\DeclareCaptionStyle{ruled}{labelfont=normalfont,labelsep=colon,strut=off} % DO NOT CHANGE THIS
\floatstyle{ruled}
\newfloat{listing}{tb}{lst}{}
\floatname{listing}{Listing}
\title{Not There Yet: Evaluating Vision Language Models in Simulating the Visual Perception of People with Low Vision}

\author {
    Rosiana Natalie,
    Wenqian Xu,
    Ruei-Che Chang,
    Rada Mihalcea,
    Anhong Guo
}
\affiliations {
    University of Michigan\\
    \{rosianan, wtxu, rueiche, mihalcea, anhong\}@umich.edu
}
\usepackage{bibentry}
\begin{document}

\maketitle

\begin{abstract}
Advances in vision language models (VLMs) have enabled the simulation of general human behavior through their reasoning and problem solving capabilities.
However, prior research has not investigated such simulation capabilities in the accessibility domain. 
In this paper, we evaluate the extent to which VLMs can simulate the vision perception of low vision individuals when interpreting images.
We first compile a benchmark dataset through a survey study with 40 low vision participants, collecting their brief and detailed vision information and both open-ended and multiple-choice image perception and recognition responses to up to 25 images. Using these responses, we construct prompts for VLMs (GPT-4o) to create simulated agents of each participant, varying the included information on vision information and example image responses. We evaluate the agreement between VLM-generated responses and participants’ original answers.
Our results indicate that VLMs tend to infer beyond the specified vision ability when given minimal prompts, resulting in low agreement (0.59).
The agreement between the agent' and participants' responses remains low when only either the vision information (0.59) or example image responses (0.59) are provided, whereas a combination of both significantly increase the agreement (0.70, p $<$ 0.0001). 
Notably, a single example combining both open-ended and multiple-choice responses, offers significant performance improvements over either alone (p $<$ 0.0001), while additional examples provided minimal benefits (p $>$ 0.05).
\end{abstract}


% \section{Introduction}

\section{Introduction}

Recent advancements in large language models (LLMs) have expanded opportunities for human-AI interaction, especially in role-playing scenarios~\cite{park2023generative, wang2024survey}, driven by improved reasoning and problem-solving capabilities~\cite{dasgupta2022language, orru2023human}. LLMs have been used for role-play across domains like gaming~\cite{wang2023voyager, xu2023exploring}, social networks~\cite{park2022social}, education~\cite{lu2024generative}, and content creation~\cite{choi2024proxona}. They are particularly useful in resource-constrained settings for tasks such as data annotation~\cite{salminen2023can}, survey generation~\cite{hamalainen2023evaluating}, and usability testing~\cite{taeb2024axnav}. 

Despite growing interest in LLM-driven role-play and simulation, their application in accessibility remains largely unexplored. This presents a compelling opportunity: in situations where directly involving people with low vision is difficult due to logistical, privacy, or resource constraints, vision language models (VLMs) may serve as a promising alternative. These models could be used to conduct initial automated accessibility evaluation of visual media, simulate and pilot first-person accounts of technology use, perform low-cost prototyping of assistive technologies, among others, to support inclusive design processes. They also hold potential for powering adaptive and personalized technologies that respond to users’ unique vision abilities. Thus, understanding how well VLMs can simulate visual perception is a critical step toward realizing this opportunity.

In this paper, we investigate the current capabilities and limitations of employing VLMs to simulate the vision abilities of people with low vision. Specifically, our study addresses: \textbf{What is the vision perception simulation performance of current VLMs? How do prompt, context, and example format affect simulation performance?}

We first collect a benchmark dataset with 40 low vision participants and gather self-reported vision information, and responses to image perception and recognition tasks using both open-ended descriptions and multiple-choice questions. From this survey, we collect 709 open-ended descriptions (avg. word count 27.2) and 4,170 multiple-choice responses. Low vision participants answer an average of 60.4\% of the multiple-choice questions correctly, with individual scores ranging from 0\% to 99\%, reflecting a wide range of visual accessibility. Participants also complete the survey in approximately 70.2 minutes on average.

Next, we use this dataset to systematically evaluate a range of VLM prompt configurations for constructing simulated agents of low vision participants. Specifically, we examine: 
\begin{itemize}
    \item \textbf{RQ1:} What is the performance of simulated agents with little to no prompting?
    \item \textbf{RQ2:} How do vision information and example image responses affect simulation performance?
    \item \textbf{RQ3:} How do the format and number of open-ended descriptions and multiple-choice responses affect simulation performance?
\end{itemize}

Our results show that VLMs tend to infer beyond the specified visual perception abilities, thus resulting in low agreement with human responses. Compared to the true answers for the images perceived by sighted people, the agent achieves an \textit{accuracy} of 0.94 with no prompting, and 0.92 with minimal prompting. These result in a low \textit{agreement} between the low vision participants and the agents, which are 0.59 (no prompting), and 0.59 (minimal prompting). On the other extreme, when prompted being blind, the agents achieve 0 accuracy (because it answers all questions with ``I Can't Tell''),  resulting in a lower agreement of 0.35. 

Furthermore, the agreement between the agents’
and participants’ responses remain low when only either the vision information (0.59) or example image responses (0.59)
are provided, whereas a combination of both significantly increase the agreement (0.70, p $<$ 0.0001). Notably, a single
example combining both open-ended and multiple-choice responses offer significant performance improvements over either alone (p $<$ 0.0001), while additional examples provided minimal benefits (p $>$ 0.05). With these agreement scores (highest mean = 0.70), we believe that VLMs' capabilities to simulate the visual perception of people with low vision is ``Not There Yet.''

Overall, our study contributes:

\begin{enumerate}
    \item A benchmark compiled from 40 people with low vision, consisting of vision information and image perception responses.

    \item The design of various prompt configurations for creating VLM-based agents that simulate people with low vision's vision perception.

    \item Insights on agent performance across different prompt configurations and how different types, formats, and the number of examples provided during prompting impacted simulation performance.
    
\end{enumerate}

\subsection{Ethical and Societal Considerations}
While simulations can offer benefits (e.g., supporting accessibility evaluation, and low-cost prototyping), they also raise important ethical concerns. Disability simulations have been critiqued for reinforcing stereotypes~\cite{Bennett2019ThePO}, trivializing lived experiences~\cite{Cossovich2023CodesigningNK}, misrepresenting disability~\cite{Morris2019AIAA}, and excluding disabled people from the design process~\cite{NarioRedmond2017CripFA}. We acknowledge these valid critiques and agree that simulating disabled experiences should be approached with caution~\cite{andrew2022accessible}. However, we argue that it is equally important to explore how emerging AI capabilities might be leveraged in ways that benefit the accessibility community, and actively involve people with disabilities in shaping future technologies. This work presents an initial evaluation of the strengths and limitations of VLM-based simulation agents -- not to suggest they replace human participants, but to understand their potential to support early-stage design when direct involvement is not feasible.
\section{Related Work}
Our work is related to the body of literature on disability simulations, and the emerging capabilities of LLMs in simulating human behavior.

\subsection{Simulating People with Disabilities}
Disability simulation refers to methods that approximate the experiences of people with disabilities, typically to foster empathy, inform design, or evaluate accessibility~\cite{Bennett2019ThePO}. Prior research has explored simulations in various disabilities, including mobility impairments~\cite{Ehibhatiomhan2022ALI}, visual impairments~\cite{juniat2019understanding, Kim2018EmpathDVE, Barbieri2023RealterAI, zhao2018enabling}, cognitive disabilities~\cite{Ehibhatiomhan2022ALI}, and hearing loss~\cite{Nelson2023SteppingIT}. 

Researchers and practitioners have explored many techniques to specifically simulate visual impairments for different purpose.
For example, for direct and physical manipulation of the vision, prior studies explore the use of techniques including simple blindfolding~\cite{Colwell2013SimulatingDA}, the use of the dedicated glasses or goggles that works as filters to mimic visual impairments experience~\cite{juniat2019understanding}. Prior research also has explored methods of simulating visual impairments in digital and immersive technologies, such as the use of VR and AR technologies which are used to create a simulation environment in virtual environment~\cite{Hkkil2018IntroducingVR, Barbieri2023RealterAI, zhao2017understanding, zhao2018enabling} and displaying the outcome of image processing system. Furthermore, Kim et al. explored the use of mobile app to display the image filter on screen~\cite{Kim2018EmpathDVE}. Lastly, several studies in the game domain have explored interactive ways in simulating visual impairments through games~\cite{Leo2024DisabilityRA, melthis2015using}. These simulations serve a range of purposes, including raising awareness among non-disabled individuals~\cite{juniat2019understanding, Leo2024DisabilityRA, melthis2015using}, healthcare and professional training~\cite{juniat2019understanding}, informing inclusive design~\cite{Kim2018EmpathDVE}, and evaluating assistive technologies~\cite{Hwang2018ImpactOO, Acevedo2022RealtimeLV, Almutleb2020TheEO, Kim2018EmpathDVE}. 

We acknowledge the risks associated with disability simulations, while also recognizing their potential value for early-stage ideation, testing, and understand the user profile as to support the effort for more personalized and adaptive experience with AI based applications.
Our work aims to extend this body of simulation research and perform an initial evaluation of the performance of VLMs for vision perception simulation.

\subsection{Role-Playing and Simulations with LLMs}
LLMs have recently emerged as powerful tools for simulating human behavior across diverse domains. They have been studied in strategic gaming~\cite{xu2023exploring}, social interaction~\cite{park2022social, park2023generative}, education~\cite{lu2024generative}, content creation~\cite{choi2024proxona}, and the social sciences~\cite{huang2024social}, where they are used to perform roles that involve reasoning, communication, and decision-making. \citeauthor{park2023generative}’s Generative Agents demonstrated how LLM-powered characters could simulate daily routines and social interactions in a virtual world~\cite{park2023generative} and replicate answer for General Social Survey~\cite{park2024generative}. Similarly, Social Simulacra used LLMs to populate prototype social computing platforms with simulated users, allowing designers to explore emergent social behaviors before deployment \cite{park2022social}. Other studies have leveraged LLMs to replicate empirical findings in social science research~\cite{huang2024social}, and generate synthetic datasets for open-ended tasks~\cite{hamalainen2023evaluating}.

Recent studies further suggest that LLMs can simulate aspects of human behavior where involving real participants may be resource-constrained or logistically challenging~\cite{liu2025free,hamalainen2023evaluating}. Researchers have studied the potential of such agents to perform labor-intensive tasks, including survey response generation~\cite{hamalainen2023evaluating}, data annotation~\cite{salminen2023can}, and early-stage user research exploration~\cite{taeb2024axnav, lu2025uxagent, liu2025free}. These applications demonstrate the growing utility of LLMs in modeling human behavior, especially when personas or prompts are carefully designed to reflect specific roles or contexts.

However, most LLM role-play applications have focused on cognitive, social, or task-based personas. There is a notable gap in exploring LLMs, particularly VLMs, as proxies for users with perceptual disabilities, such as visual impairments. Despite growing interests in applying AI to accessibility, little work has examined whether VLMs can meaningfully simulate the visual perception of people with visual impairments. Our work addresses this gap by investigating the capabilities and limitations of current VLMs in simulating visual perception, and examining how different input information in prompts influences the quality of simulations.

A key insight from prior research is that personas become more believable and useful when enriched with contextual and personality traits~\cite{shanahan2023role}. Building on this, we extend prior approaches by grounding our VLM-based agents prompts in real-world data from individuals with low vision. Inspired by works such as Proxona~\cite{choi2024proxona}, Park et al. \cite{park2024generative}, Shin et al.~\cite{shin2024understanding}, and Hämäläinen et al.~\cite{hamalainen2023evaluating}, which use real participant data to build contextually rich personas, we embed actual survey-derived descriptions of visual conditions and examples of their image recognition and perception into the prompts. 
\section{Benchmark of Human Vision Information and Image Perception}

Our benchmark consists of vision information as well as vision perception and recognition tasks from 40 low vision participants. We recruit participants through the National Federation of the Blind (NFB) research participant solicitation request form, university mailing lists, and word of mouth. To be eligible, individuals had to be at least 18 years old and self-identify as having low vision, with some degree of light perception and usable vision.

To compile a benchmark of vision information and image perception from low vision individuals, we develop a custom survey web interface, which is screen reader-compatible, and support image enlarging and minimizing). We log participants' change of responses and the duration participants spent on each question. The survey is designed to take approximately one hour to complete, and participants receive a \$20 incentive for participating. The study protocol was approved by the Institutional Review Board (IRB) at our university.

\subsection{Survey Questions}
The survey is structured into three main sections:

\subsubsection{Brief Vision Information} We ask participants about their vision-related background, including their level of vision (e.g., visual acuity, visual field, and color perception), the onset of their vision impairment (i.e., congenital or acquired), their vision progression over time, and medical diagnoses.

\subsubsection{Detailed Vision Information} We ask participants to provide more detailed descriptions of how they perceive their surroundings (e.g., shapes, colors, motion, and variations under different lighting conditions). We also ask them to describe any unique aspects of their visual experience that might differ from others with the same medical diagnosis. Capturing these individual differences helps move beyond stereotypical assumptions tied to specific diagnoses and gather more personalized understanding of vision information. We also include a subset of items from the Visual Functioning Questionnaire (VFQ-25)~\cite{pawar2021assessment}, a widely used tool for assessing the impact of vision on daily living. Given VFQ-25's breadth, we only select items that relate to participants’ ability to perceive actions, objects, and facial expressions.

\subsubsection{Image Perception and Recognition}
To capture how participants use their vision to perceive images, we curate a list of 25 images. 
They are selected following the classification in~\citeauthor{stangl2021going} to represent a range of visual elements and contexts, drawing from the intersection of five content types, including
(1) politician/people, 
(2) living room, 
(3) bazaar, 
(4) mountains, and 
(5) food; 
as well as five usage scenarios, including 
(i) news for learning, 
(ii) e-commerce for purchasing, 
(iii) social networking for information seeking, 
(iv) travel for planning, and 
(v) library for knowledge sharing.

We ask participants both open-ended and multiple-choice questions. Participants first respond to the open-ended question: \textit{``Based on your visual perception, how would you describe this image? Please describe any shapes, colors, details, or other elements that you can perceive. If certain aspects are unclear or not visible to you, feel free to describe how you perceive the image in your own way.''}

\begin{figure}
    \centering
    \includegraphics[width=\linewidth]{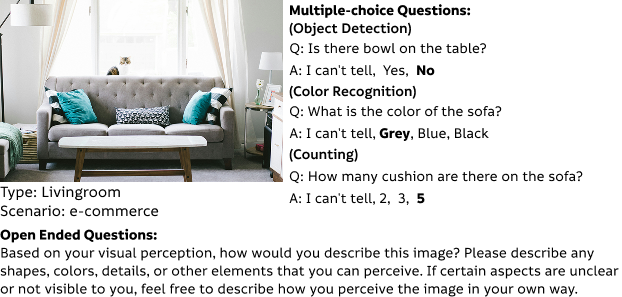}
        \caption{Sample image and corresponding questions with the living room type and the e-commerce for purchasing scenario. The bolded answers indicate the correct ones.}
    \vspace{-1.2pc}
    \label{fig:survey_questions}
\end{figure}

Next, participants answer six multiple-choice questions (MCQs) intended to cover three visual skills suggested by prior work~\cite{zeng2020vision}: (i) object recognition (\textit{e.g.,} What is the person wearing?), (ii) color recognition (e.g., What is the color of the person’s hair?), and (iii) counting (e.g., What is the total count of people in the image?). An example image along with its corresponding MCQs is shown in Figure~\ref{fig:survey_questions}. These structured questions provide a consistent basis for assessing visual perception across participants and are later used as the ground truth of the VLM-based agent’s performance. We iteratively refine the questions to be clear and answerable, and two sighted individuals both achieve 100\% accuracy on the final set. See Appendix A for an overview of the survey interface, Appendix B for the task images, and Appendix C for the full list of survey questions.

\subsection{Benchmark Statistics}
We collect survey responses from 40 low vision participants (age: M = 48 years, SD = 18.0). Participants reported using various assistive technologies to complete the survey, including screen magnifiers (N = 14), screen readers (N = 19), magnifying lenses (N = 8), and braille readers (N = 2).
On average, participants spent approximately 70.2 minutes (SD = 32.3) to complete the survey. Due to the diverse vision ability and assistive tool use that affected task completion time, we made the first 10 images mandatory. For those 10, participants took an average of 29.5 minutes (SD = 21.3). Within the one-hour mark, participants completed an average of 16 questions.

Among the participants, 9 complete all 25 images, while 8 complete only the 10 mandatory images. In total, we collect 4,170 MCQ responses and 709 open-ended descriptions from participants, with an average word count per description of 27.2 (SD=31.6). Breaking down the duration for each survey section, participants spend an average of 5.2 minutes (SD = 5.6) on the Brief Vision Information section, 8.8 minutes (SD = 5.8) on the Detailed Vision Information section, 1.8 minutes (SD = 3.3) on each open-ended image description, and 1.0 minute (SD = 0.8) on each group of MCQs.

\begin{figure}
    % \centering
    \includegraphics[width=\linewidth]{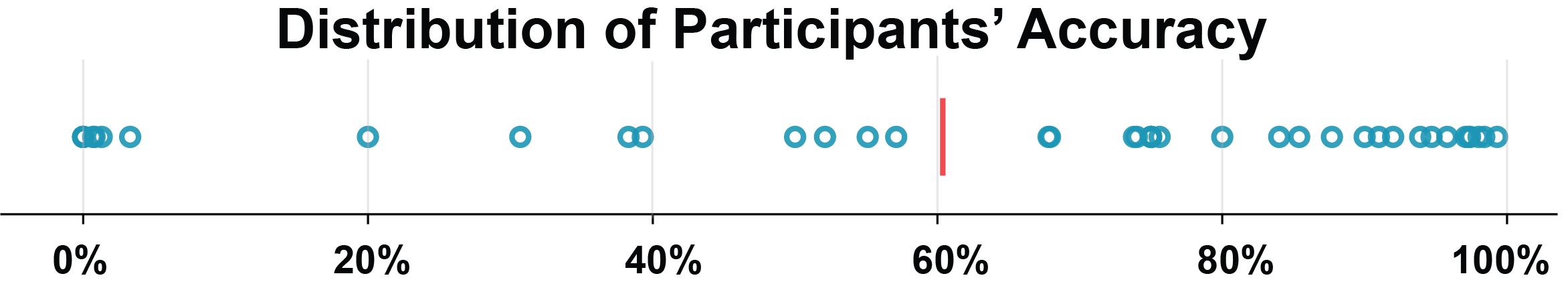}
        \caption{The accuracy distribution of participants' answers compared to the true answers perceived by sighted people. The mean accuracy is 60.4\% (SD = 36\%), shown in red vertical line, indicating a diverse range of vision abilities. }
    \vspace{-1.2pc}
    \label{fig:participant_performance}
\end{figure}

We calculate the percentage of questions participants answer correctly, as a measure of task difficulty due to their vision abilities. Participants on average answer 60.4\% (Std = 36\%, Md = 74.5\%) of the MCQs correctly, with individual percentages ranging from 0\% to 99\%, showing a wide range of visual abilities (Figure~\ref{fig:participant_performance}).

\section{Evaluating VLM Simulation Performance}
We evaluate VLM simulation performance by designing baseline prompts with no or minimal vision information and prompts that incorporate participants' survey responses. Performance is measured using agreement scores between the agents' responses and participants' answers.

\subsection{Prompt Design}
We first design prompts to predict answers with no or minimal vision information about the participants the baselines. Then, we constructed agents based on their survey responses, aiming to reflect participants' unique visual abilities. We illustrated the prompt template in Appendix D.

For the baseline prompts, we used prompts with no or minimal vision information about the participants. We prompt VLMs to be one of three roles: an assistant with full vision (i.e., a sighted agent), an assistant with no visual perception (i.e., a blind agent), and an assistant with unspecified visual impairments (i.e., a minimally-prompted agent).

Then, we design agent prompt that include the responses from the participants. We design in total 16 prompts (Appendix E) that include either vision information, image perception and recognition responses (example), or both.

\subsubsection{Vision Information} 
We vary three levels of vision information detail:
    
\begin{itemize}

    \item \textbf{Diagnosis-only (diagnosis):} The prompt includes a brief medical diagnostic statement reported by the participant in the Brief Vision Information survey section (e.g., Retinitis Pigmentosa, Glaucoma).
    
    \item \textbf{Brief Vision information (brief):} The prompt includes the participant’s response to the Brief Vision Information section of the survey.
    
    \item \textbf{Detailed Vision information (detailed):} The prompt incorporates the participant’s response from Detailed Vision Information Section.
\end{itemize}

\subsubsection{Example} 
Building on prior work that highlights the benefits of including explicit examples to guide model responses ~\cite{lu2024generative,brown2020language}, we prompt the VLM using only examples and combining vision information with examples. We vary the examples along three dimensions:

\begin{itemize}
     \item \textbf{Number of examples:} single or multiple.
     
     \item \textbf{Type of single example:} examples drawn from (i) unrelated images, (ii) the same image type, or (iii) the same image scenario.
    
    \item \textbf{Type of example response:} open-ended responses, MCQ responses, or both.
    
\end{itemize}

For the single-example design, agents are given one example answer drawn from one of three types: (i) an image of the same type, (ii) an image from the same scenario, or (iii) an image unrelated in type or scenario. For each type, we also design prompts that include either only the open-ended responses, only the MCQ responses, or both.

For the multiple-example design, agents are given nine example answers corresponding to all mandatory images except the one being used for prediction (i.e., a leave-one-out setup). Similar to the single-example design, we designed prompts that included either only the open-ended response, only MCQ responses, or both.

\subsubsection{Prediction}For all prompt variations, the agent is instructed to predict open-ended and MCQ responses from the image perception and recognition task. For all prompts requiring examples, we fine-tune and predict using responses to the participant's mandatory images (i.e., the ten images covering combinations of image types and scenarios).

\subsection{Implementation}
We use OpenAI’s GPT-4o (version: gpt-4o-2024-11-20), a large multimodal model capable of processing both text and image queries.
For each API call, the model is prompted to generate a response to a single image-based perception or recognition question.
We set the model’s temperature to 0.

\subsection{Metrics}

We evaluate the agent’s performance on the MCQ responses by calculating the \textit{agreement} score, the proportion of questions where both the agent and the participant provided the same answer. The \textit{agreement} ranges from 0 (no agreement) to 1 (perfect agreement).

We use a Generalized Linear Mixed Model (GLMM) with a binomial distribution and a logit link function. The response variable is agreement (i.e., a binary indicator of whether the agent agrees or disagrees with the participant). Prompt design is treated as a fixed effect, and both participant and questions are treated as random effects. This model estimate the probability of agreement between the agent and the participant under each prompt design. We then fit our GLMM models to assess the effects of different prompt designs and conduct pairwise comparisons using Estimated Marginal Means (EMMs) to test for differences between conditions.
\section{Evaluation Results}

\subsection{RQ1: What is the performance of simulated agents with little to no prompting?} \label{sec:raw_performance}

Our results indicate that VLMs tend to infer beyond the specified vision ability when given minimal prompts. Agents created with no to minimal prompt are achieving high accuracy against the ground truth in general, which are 0.94 for  sighted and 0.92 for minimally-prompted agent. Blind agents answer all questions \textit{``I can't tell''}, which results in 0 accuracy. This is intuitive because we did not pass them any visual image, simulating that they have no vision perception.

This high accuracy of naive prompting compared to true answers perceived by sighted people results in low agreement between the agents and the participants (i.e., sighted agent: Mean = 0.59, SD = 0.33; minimally-prompted agent: Mean = 0.59, SD = 0.32). Moreover, the agreement is even lower between participants and the blind agent (Mean = 0.35, SD = 0.36).

\begin{figure}
    \centering
    \includegraphics[width=\linewidth]{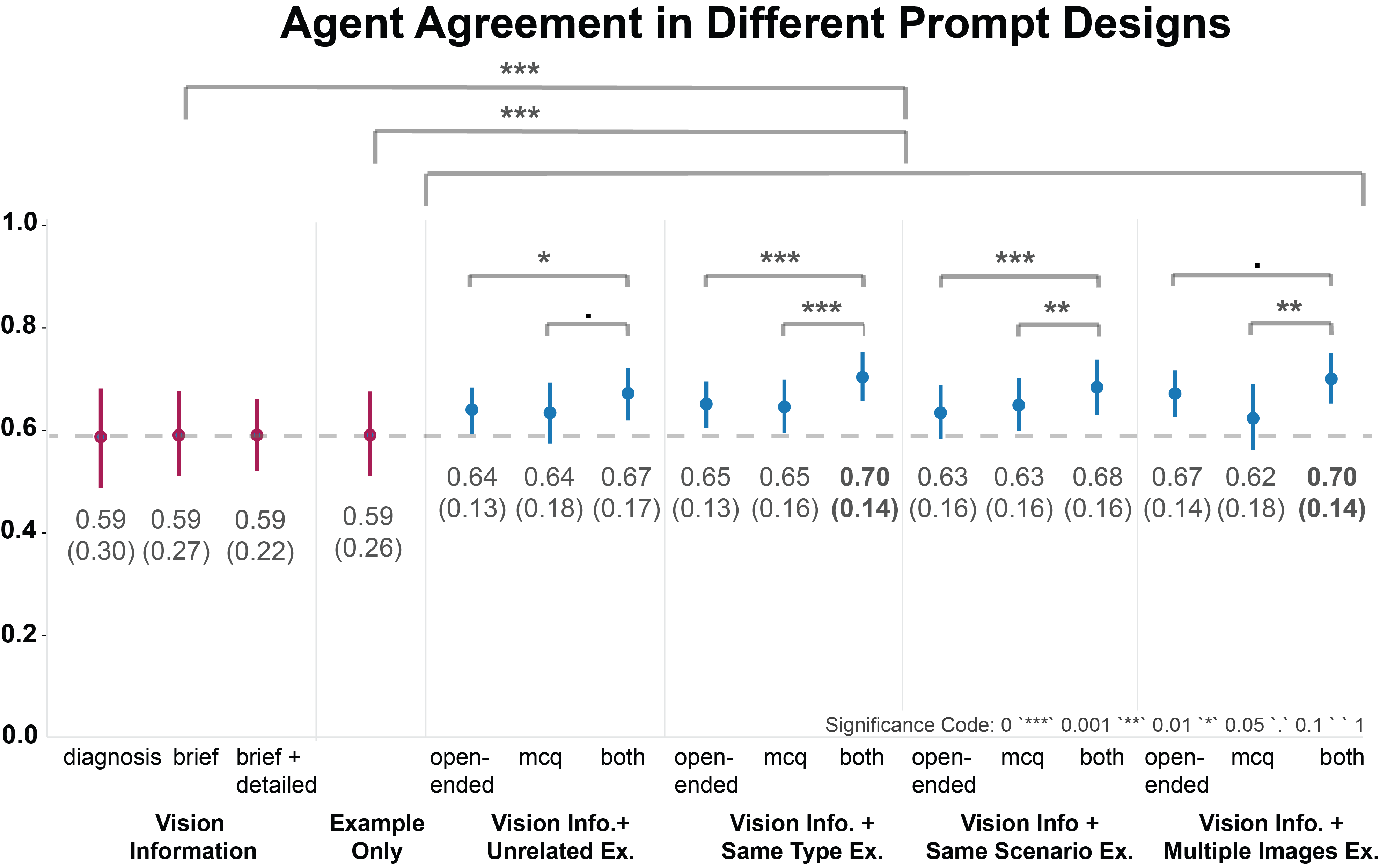}
    \caption{The point plot shown for different performance for different prompt feedback. The prompts are either using vision profile only or with the example. The graph shows increasing trend between the visual profile and examples, and within the question types, open-ended, multi-choice questions, or both, also shows increasing trend. We present the mean and standard deviation value (in bracket).}
    \vspace{-1pc}
    \label{fig:condition_performance}
\end{figure}

\subsection{RQ2: How do vision information and example image responses affect simulation performance?} \label{sec:result_info}

\textbf{Vision information.} When prompting with more vision information, the agreement between the agent's and participants' responses remains low when only the vision information. The average agreement scores are 0.59 (SD = 0.30) for the prompt with diagnosis, 0.59 (SD = 0.27) for the prompt with brief vision description, and 0.59 (SD = 0.22) for the prompt brief and detailed description.

We find no significant difference in agreement across the vision information-only prompt designs. Pairwise comparisons reveal no statistically significant differences between the prompt designs, where all $p$ $>$ 0.1 (diagnosis vs. brief: $z$ = -1.01, $p$ $>$ 0.1; diagnosis vs. detailed: $z$ = -1.66, $p$ $>$ 0.1; brief vs. diagnosis: $z$ = -0.65, $p$ $>$ 0.1).

\textbf{Example.} Agreement improves when examples are included in the prompts. Specifically, mean agreement scores increase to 0.65 (SD = 0.160) with an unrelated image example, 0.67 (SD = 0.15) with an image of the same type, 0.66 (SD = 0.16) with a same-scenario image, and 0.67 when multiple examples are provided. All example-based conditions show significantly higher agreement compared to the vision information-only prompts (vision information vs. unrelated: $z$ = 10.43, $p$ $<$ 0.0001; vs. same type: $z$ = 12.71, $p$ $<$ 0.0001; vs. same scenario: $z$ = 11.15, $p$ $<$ 0.0001; vs. multiple examples: $z$ = 12.81, $p$ $<$ 0.0001).

We also run additional analyses to evaluate the outcome of prompts where we provided only examples without any vision information. The mean agreement is 0.59 (SD = 0.25), which is significantly lower than the agreement achieved when both vision profiles and examples are combined (example only vs. example + vision information: $z$ = -8.525, $p$ $<$ 0.0001).
 
Upon manual inspection of the agent responses for open-ended tasks, the agent often produce surprisingly detailed and comprehensive image descriptions -- even when prompted with information indicating vision impairment. While the agent occasionally acknowledge its visual limitations with statements such as, \textit{``I'm unable to identify specific details or people in the image, but I can describe general elements''}, these acknowledgments are frequently followed by unexpectedly specific and confident responses. For example, one response states:
\textit{``It seems like a top-down view of a table setting. I can perceive round shapes that might be plates, and there are some contrasting colors that could be food items. The background has a pattern that might be tiles. There are also some elongated shapes that could be utensils or hands. The overall scene gives a sense of a meal or gathering.''}, which appears implausibly detailed for someone who could not identify specific details of objects or people.

In contrast, when the prompts include example answers from participants for other images, the agent’s responses displayed greater uncertainty and caution. Phrases such as \textit{``vague...''}, \textit{``possibly...''}, or \textit{``it seems like...''} are more frequent. In some cases, the agent explicitly states its limitations, responding with phrases like \textit{``I can’t tell''} or \textit{``Unclear''}, particularly when the provided vision information described a high degree of visual impairment.
 
These findings suggest that vision information alone is insufficient to constrain the agent’s outputs in a way that reflected the intended perceptual abilities. In contrast, adding examples to the vision information in the prompts help guide the agent toward more appropriately accurate responses.

\subsection{RQ3: How do the format and number of open-ended descriptions and multiple-choice responses affect simulation performance?}

\textbf{Example format.} We investigate which types of example-based prompts lead to better agreement. Notably, examples that combine both open-ended and multiple-choice responses offer significant performance improvements over either format alone.

Across all prompts that incorporate examples, we find that including both open-ended description responses and MCQ responses in the examples results in the highest agreement scores. Specifically, the average agreement is 0.67 (SD = 0.17) for examples from unrelated images, 0.70 (SD = 0.14) for same-type images, 0.68 (SD = 0.16) for same-scenario images, and 0.70 (SD = 0.14) for prompts with multiple examples—all of which include both open-ended and MCQ responses.

These scores are significantly higher than those from prompts that included only open-ended response examples (unrelated: Mean = 0.64, SD = 0.13 ($z$ = 2.51, $p$ $<$ 0.05); same type: Mean = 0.65, SD = 0.13 ($z$ = 4.27, $p$ $<$ 0.0001); same scenario: Mean = 0.63, SD = 0.16 ($z$ = 4.01, $p$ $<$ 0.0001); multiple: Mean = 0.67, SD = 0.14 ($z$ = 2.51, $p$ $<$ 0.05)) or only MCQ response examples (unrelated: Mean = 0.64, SD = 0.18 ($z$ = 2.51, $p$ $<$ 0.01); same type: Mean = 0.65, SD = 0.16 ($z$ = 4.73, $p$ $<$ 0.0001); same scenario: Mean = 0.65, SD = 0.16 ($z$ = 2.90, $p$ $<$ 0.01); multiple: Mean = 0.62, SD = 0.18 ($z$ = 3.04, $p$ $<$ 0.01)).

\textbf{Number of examples.}
Providing more examples offers minimal additional benefits compared to using only one example image response. Among all prompts that include examples, we find no statistically significant differences in pairwise comparison analyses between the prompt with different numbers of examples (single unrelated vs. multiple: $z$ = 2.33, $p$ $>$ 0.1; single same type vs. multiple: $z$ = 0.07, $p$ $>$ 0.1; single same scenario vs. multiple: $z$ = 1.60, $p$ $>$ 0.1).

 \section{Discussion and Future Work}
Previous studies have explored the ongoing challenges in using VLMs to simulate human behavior, such as the difficulty in making these models effectively ``unlearn'' previously acquired abilities~\cite{lu2024generative}, including, in our case, visual sensory perception. For example, our observations indicate that the model struggle to authentically simulate the experience of low vision, even when explicitly instructed to role-play as an individual with a specific vision information (Section~\ref{sec:raw_performance}). However, we identify a promising direction: incorporating few-shot examples that demonstrate how a participant with low vision performs specific tasks (e.g., image perception and recognition) can significantly enhance the fidelity of VLM-based simulated agents.

In the following sections, we discuss simulation readiness, potential applications of simulated agents with low vision, and the need for vision information datasets supported by cost-effective and generalizable data collection pipelines.

\subsection{Simulation Readiness: Limitations, Risks, and Responsible Use}

This study investigates how well VLMs can simulate the visual perception of individuals with low vision. While our findings show approximately 70\% agreement between simulated agents and actual user responses, VLM-based simulations may not yet be ready for standalone deployment and direct decision-making. Their use should be guided by caution, critical reflection, and human-in-the-loop validation. The divergence between agent outputs and human responses also amplifies a set of existing underlying deployment risks. Hallucinated or inaccurate descriptions~\cite{chen2025vision, bai2025hallucination}, biases in training data, poor generalization across diverse environments exclusionary ~\cite{zhao2024vialm, buffalo2024assistive} may produce stereotyped or culturally insensitive representations~\cite{hali2022bias}. These risks may misinform how blind and low vision (BLV) users are represented and understood in design processes. Mistrust may emerge when users encounter unpredictable failures, especially if system limitations are not communicated clearly~\cite{ahmadi2024risks}. Together, these findings underscore that VLM-based simulation should not be treated as a substitute for direct user research, but as a complementary and contingent tool whose limitations must be made explicit. Future work should focus on defining acceptable fidelity thresholds -- what level of agreement is ``good enough'' for various design contexts -- and establishing those thresholds through collaboration with BLV individuals. Understanding how BLV users perceive and consent to the simulation of their experiences is also important to ensure transparency, and trust throughout the design process.

\subsection{Broader Applications of Simulated Agents with Low Vision}

In this paper, we focus on the data collection and evaluation of a variety of prompting strategies to explore the effectiveness of simulated agents for low-vision individuals. 
Drawing from prior work in applying simulated agents to practical applications \textit{e.g.,}~\cite{hamalainen2023evaluating,park2024generative}, we also exemplify several potential use cases of our findings in this section. 
Firstly, commercial VLM-powered applications (e.g., SeeingAI~\cite{seeingai}, Be My AI~\cite{bemyai}) enable BLV users to access visual information by capturing photos and querying the application. 
However, the generalized, one-size-fits-all information typically produced by VLMs may not align with individual-specific needs, causing additional interaction such as multiple queries to obtain precise information~\cite{stangl2021going}. 
To mitigate this challenge, application developers could adopt our prompt strategies to simulate diverse low-vision user experiences, ensuring that the generated descriptions align better with users' existing comprehension. 
For instance, the agent could indicate what users can or cannot perceive from the images, which could be utilized to tailor the feedback and information provided by such applications.

Second, following an approach similar to~\cite{taeb2024axnav}, web-based systems (e.g., social media platforms) could use low-vision agents to run in the background to continuously evaluate image accessibility. 
These agents could automatically examine accessibility problems and apply corresponding visual filters in real time. For example, if an image includes color combinations that are challenging for users with color vision deficiencies (e.g., red-green colorblindness), a filter such as Daltonization~\cite{daltonization} could dynamically adjust the colors specific to user needs, thus enhancing image accessibility.

Future research can expand on these use cases by creating detailed profiles of users' vision capabilities to develop representative simulated agents by referring to our prompting strategies. 
These agents can then proactively identify and address accessibility issues, refining solutions before their deployment to actual users.

\subsection{Cost-Effective and Generalizable Benchmark Data Collection Pipeline}

This study introduces a novel, generalizable approach to collecting structured vision information for simulating agents representing diverse low-vision conditions. Our findings show that combining self-reported vision data with targeted image perception tasks and prompting strategies played a critical role in helping VLMs approximate the perception of individuals with low vision (Section~\ref{sec:result_info}). However, collecting accurate and diverse vision profiles remains challenging due to the absence of standardized, scalable methodologies. Traditional clinical tools like the Snellen Chart~\cite{snellen} and Amsler Grid~\cite{amsler}, while valuable~\cite{wang2024low}, often lack accessibility and scalability.

To address this, we use an online survey paired with structured visual tasks to gather individual vision data. This method proves both efficient and accessible. For example, collecting the minimal data needed for simulation (i.e., vision information and an image example response) takes only about 15 minutes, and within an hour, participants can respond to 16 image-based tasks—allowing for a broader range of image types to be included when needed. Importantly, participants report that the survey help them reflect on their own vision capabilities, suggesting that the data collection process itself can offer personal value. 

Future work can develop end-to-end pipelines for collecting richer, longitudinal vision data, potentially through apps or in-situ studies, and by exploring ways to integrate simulation into real-world tools that support both user reflection and data-driven design feedback.

\subsection{Potential for Improved Performance with Advanced Models}
As language models continue to evolve, so does the potential to enhance the accuracy and reliability of the simulations. While our evaluation is currently limited to a non-reasoning model due to the high cost and restricted access to advanced reasoning models (e.g., GPT-o3-pro), our initial results still suggest promising potential of VLM agents in simulating the visual perception of individuals with low vision. 

Advances in model capabilities are likely to improve consistency in visual reasoning and enhance the reliability of simulated responses. As more powerful models become increasingly affordable and accessible, we anticipate continued improvements in simulation quality, paving the way for more effective, scalable tools for early-stage accessibility evaluation and personalized user experiences.

\section{Conclusion}

Our study investigates the use of VLMs to simulate the visual perception of individuals with low vision. By gathering a structured benchmarking dataset from 40 participants and conducting a series of prompting, we assesses the performance of the agent and how different types and amounts of input information influence simulation performance.

Our findings show that VLMs often infer beyond the intended visual constraints without sufficient prompting. However, providing a single example that includes both open-ended and MCQs responses significantly improves alignment, while additional examples offer minimal gains.

This study evaluates the readiness of VLMs to simulate visual perception of people with low vision, finding that while simulations show promise, fidelity gaps may add to ongoing risks (e.g., hallucinations and bias). Such simulations should be used as complementary tools, supported by human-in-the-loop validation and clearly defined fidelity thresholds. Looking ahead, simulated agents may support broader applications, from personalized assistive technologies to automated accessibility auditing.

\bibliography{main.bib}

\begin{thebibliography}{53}
\providecommand{\natexlab}[1]{#1}

\bibitem[{Acevedo et~al.(2022)Acevedo, Colantoni, Dinet, and Tr{\'e}meau}]{Acevedo2022RealtimeLV}
Acevedo, V.; Colantoni, P.; Dinet, {\'E}.; and Tr{\'e}meau, A. 2022.
\newblock Real-time Low Vision Simulation in Mixed Reality.
\newblock \emph{2022 16th International Conference on Signal-Image Technology \& Internet-Based Systems (SITIS)}, 354--361.

\bibitem[{Ahmadi and Lewis(2024)}]{ahmadi2024risks}
Ahmadi, N.; and Lewis, J. 2024.
\newblock Reporting Risks in AI-based Assistive Technology Research: A Systematic Review.
\newblock \emph{arXiv preprint arXiv:2407.12035}.

\bibitem[{Almutleb and Hassan(2020)}]{Almutleb2020TheEO}
Almutleb, E.~S.; and Hassan, S.~E. 2020.
\newblock The Effect of Simulated Central Field Loss on Street-crossing Decision-Making in Young Adult Pedestrians.
\newblock \emph{Optometry and Vision Science}, 97: 229 -- 238.

\bibitem[{Andrew and Tigwell(2022)}]{andrew2022accessible}
Andrew, S.; and Tigwell, G.~W. 2022.
\newblock Accessible design is mediated by job support structures and knowledge gained through design career pathways.
\newblock \emph{Proceedings of the ACM on Human-Computer Interaction}, 6(CSCW2): 1--24.

\bibitem[{Azzam and Ronquillo(2023)}]{snellen}
Azzam, D.; and Ronquillo, Y. 2023.
\newblock Snellen chart.
\newblock In \emph{StatPearls [Internet]}. StatPearls Publishing.

\bibitem[{Bai et~al.(2025)Bai, Wang, Zhao, and Liu}]{bai2025hallucination}
Bai, Y.; Wang, X.; Zhao, M.; and Liu, Q. 2025.
\newblock Hallucination of Multimodal Large Language Models: A Survey.
\newblock \emph{arXiv preprint arXiv:2503.00029}.

\bibitem[{Barbieri et~al.(2023)Barbieri, Albanese, Capris, Canessa, Sabatini, and Sandini}]{Barbieri2023RealterAI}
Barbieri, M.; Albanese, G.~A.; Capris, E.; Canessa, A.; Sabatini, S.~P.; and Sandini, G. 2023.
\newblock Realter: An Immersive Simulator to Support Low-Vision Rehabilitation.
\newblock In \emph{XR}.

\bibitem[{BeMyAI(2025)}]{bemyai}
BeMyAI. 2025.
\newblock Introducing Be My AI (formerly Virtual Volunteer) for People who are Blind or Have Low Vision, Powered by OpenAI’s GPT-4.

\bibitem[{Bennett and Rosner(2019)}]{Bennett2019ThePO}
Bennett, C.~L.; and Rosner, D.~K. 2019.
\newblock The Promise of Empathy: Design, Disability, and Knowing the "Other".
\newblock \emph{Proceedings of the 2019 CHI Conference on Human Factors in Computing Systems}.

\bibitem[{Brown et~al.(2020)Brown, Mann, Ryder, Subbiah, Kaplan, Dhariwal, Neelakantan, Shyam, Sastry, Askell et~al.}]{brown2020language}
Brown, T.; Mann, B.; Ryder, N.; Subbiah, M.; Kaplan, J.~D.; Dhariwal, P.; Neelakantan, A.; Shyam, P.; Sastry, G.; Askell, A.; et~al. 2020.
\newblock Language models are few-shot learners.
\newblock \emph{Advances in neural information processing systems}, 33: 1877--1901.

\bibitem[{Chen et~al.(2025)Chen, Yang, Wu, and Wang}]{chen2025vision}
Chen, Z.; Yang, H.; Wu, L.; and Wang, Y. 2025.
\newblock A Large Vision-Language Model Based Environment Perception System for Visually Impaired People.
\newblock \emph{arXiv preprint arXiv:2504.18027}.

\bibitem[{Choi et~al.(2024)Choi, Kang, Choi, Lee, and Kim}]{choi2024proxona}
Choi, Y.; Kang, E.~J.; Choi, S.; Lee, M.~K.; and Kim, J. 2024.
\newblock Proxona: Leveraging LLM-Driven Personas to Enhance Creators' Understanding of Their Audience.
\newblock \emph{arXiv preprint arXiv:2408.10937}.

\bibitem[{Colwell(2013)}]{Colwell2013SimulatingDA}
Colwell, C.~M. 2013.
\newblock Simulating disabilities as a tool for altering individual perceptions of working with children with special needs.
\newblock \emph{International Journal of Music Education}, 31: 68 -- 77.

\bibitem[{Cossovich et~al.(2023)Cossovich, Hodges, Kang, and Girouard}]{Cossovich2023CodesigningNK}
Cossovich, R.; Hodges, S.; Kang, J.; and Girouard, A. 2023.
\newblock Co-designing new keyboard and mouse solutions with people living with motor impairments.
\newblock \emph{Proceedings of the 25th International ACM SIGACCESS Conference on Computers and Accessibility}.

\bibitem[{Dasgupta et~al.(2022)Dasgupta, Lampinen, Chan, Sheahan, Creswell, Kumaran, McClelland, and Hill}]{dasgupta2022language}
Dasgupta, I.; Lampinen, A.~K.; Chan, S.~C.; Sheahan, H.~R.; Creswell, A.; Kumaran, D.; McClelland, J.~L.; and Hill, F. 2022.
\newblock Language models show human-like content effects on reasoning tasks.
\newblock \emph{arXiv preprint arXiv:2207.07051}.

\bibitem[{Ehibhatiomhan et~al.(2022)Ehibhatiomhan, Foreman, Barrott, Shek, and Nabhani-Gebara}]{Ehibhatiomhan2022ALI}
Ehibhatiomhan, R.; Foreman, E.; Barrott, L.; Shek, J.; and Nabhani-Gebara, S. 2022.
\newblock ‘A life in a day’ simulation experience: perceptions of oncology nurses and pharmacy staff.
\newblock \emph{BMC Nursing}, 21.

\bibitem[{H{\"a}kkil{\"a} et~al.(2018)H{\"a}kkil{\"a}, Colley, V{\"a}yrynen, and Yliharju}]{Hkkil2018IntroducingVR}
H{\"a}kkil{\"a}, J.; Colley, A.; V{\"a}yrynen, J.; and Yliharju, A.-J. 2018.
\newblock Introducing Virtual Reality Technologies to Design Education.
\newblock \emph{Seminar.net}.

\bibitem[{Hali, Diagne, and Walker(2022)}]{hali2022bias}
Hali, M.; Diagne, B.; and Walker, J. 2022.
\newblock Measuring Representational Harms in Image Captioning.
\newblock In \emph{Proceedings of the 2022 ACM Conference on Fairness, Accountability, and Transparency (FAccT)}, 943--954.

\bibitem[{H{\"a}m{\"a}l{\"a}inen, Tavast, and Kunnari(2023)}]{hamalainen2023evaluating}
H{\"a}m{\"a}l{\"a}inen, P.; Tavast, M.; and Kunnari, A. 2023.
\newblock Evaluating large language models in generating synthetic hci research data: a case study.
\newblock In \emph{Proceedings of the 2023 CHI Conference on Human Factors in Computing Systems}, 1--19.

\bibitem[{Huang et~al.(2024)Huang, Yuan, Zhou, Guo, Wang, Zhuang, Sun, Sun, Wang, Ye et~al.}]{huang2024social}
Huang, Y.; Yuan, Z.; Zhou, Y.; Guo, K.; Wang, X.; Zhuang, H.; Sun, W.; Sun, L.; Wang, J.; Ye, Y.; et~al. 2024.
\newblock Social Science Meets LLMs: How Reliable Are Large Language Models in Social Simulations?
\newblock \emph{arXiv preprint arXiv:2410.23426}.

\bibitem[{Hwang et~al.(2018)Hwang, Tuccar-Burak, Goldstein, and Peli}]{Hwang2018ImpactOO}
Hwang, A.~D.; Tuccar-Burak, M.; Goldstein, R.~B.; and Peli, E. 2018.
\newblock Impact of Oncoming Headlight Glare With Cataracts: A Pilot Study.
\newblock \emph{Frontiers in Psychology}, 9.

\bibitem[{Juniat et~al.(2019)Juniat, Bourkiza, Das, Das-Bhaumik, Founti, Yeo, Mathew, and Okhravi}]{juniat2019understanding}
Juniat, V.; Bourkiza, R.; Das, A.; Das-Bhaumik, R.; Founti, P.; Yeo, C.; Mathew, R.; and Okhravi, N. 2019.
\newblock Understanding visual impairment and its impact on patients: a simulation-based training in undergraduate medical education.
\newblock \emph{Journal of medical education and curricular development}, 6: 2382120519843854.

\bibitem[{Kim et~al.(2018)Kim, Choo, Lee, Misra, and Balan}]{Kim2018EmpathDVE}
Kim, W.; Choo, K. T.~W.; Lee, Y.; Misra, A.; and Balan, R.~K. 2018.
\newblock Empath-D: VR-based Empathetic App Design for Accessibility.
\newblock \emph{Proceedings of the 16th Annual International Conference on Mobile Systems, Applications, and Services}.

\bibitem[{Le{\~a}o et~al.(2024)Le{\~a}o, de~Souza, Souza, Hauache, Stival, and Silla}]{Leo2024DisabilityRA}
Le{\~a}o, L.; de~Souza, D.~C.; Souza, T.; Hauache, A.; Stival, E.; and Silla, C.~N. 2024.
\newblock Disability Racer: A Digital Game for Raising Awareness of Ophthalmological: Related Issues.
\newblock In \emph{International Conference on Computer Supported Education}.

\bibitem[{Lillo et~al.(2022)Lillo, Moreira, Abad, and {\'A}lvaro}]{daltonization}
Lillo, J.; Moreira, H.; Abad, L.; and {\'A}lvaro, L. 2022.
\newblock Daltonization or colour enhancement: potential uses and limitations.
\newblock \emph{Optics Express}, 30(25): 45156--45177.

\bibitem[{Liu et~al.(2025)Liu, Sabour, Wang, and Mihalcea}]{liu2025free}
Liu, S.; Sabour, S.; Wang, X.; and Mihalcea, R. 2025.
\newblock Free Lunch for User Experience: Crowdsourcing Agents for Scalable User Studies.
\newblock \emph{arXiv preprint arXiv:2505.22981}.

\bibitem[{Lu and Wang(2024)}]{lu2024generative}
Lu, X.; and Wang, X. 2024.
\newblock Generative students: Using llm-simulated student profiles to support question item evaluation.
\newblock In \emph{Proceedings of the Eleventh ACM Conference on Learning@ Scale}, 16--27.

\bibitem[{Lu et~al.(2025)Lu, Yao, Gu, Huang, Wang, Li, Gesi, He, Li, and Wang}]{lu2025uxagent}
Lu, Y.; Yao, B.; Gu, H.; Huang, J.; Wang, J.; Li, L.; Gesi, J.; He, Q.; Li, T. J.-J.; and Wang, D. 2025.
\newblock UXAgent: An LLM Agent-Based Usability Testing Framework for Web Design.
\newblock \emph{arXiv preprint arXiv:2502.12561}.

\bibitem[{Melthis et~al.(2015)Melthis, Brown, Tang, and Hanneghan}]{melthis2015using}
Melthis, J.; Brown, A.; Tang, S.; and Hanneghan, M. 2015.
\newblock Using Serious Games to Create Awareness on Visual Impairments.
\newblock In \emph{2015 International Conference on Developments of E-Systems Engineering (DeSE)}, 165--170.

\bibitem[{Morris.(2019)}]{Morris2019AIAA}
Morris., M. G.~R. 2019.
\newblock AI and accessibility.
\newblock \emph{Communications of the ACM}, 63: 35 -- 37.

\bibitem[{Nario-Redmond, Gospodinov, and Cobb(2017)}]{NarioRedmond2017CripFA}
Nario-Redmond, M.~R.; Gospodinov, D.; and Cobb, A. 2017.
\newblock Crip for a Day: The Unintended Negative Consequences of Disability Simulations.
\newblock \emph{Rehabilitation Psychology}, 62: 324–333.

\bibitem[{Nelson, Spence, and Gormley(2023)}]{Nelson2023SteppingIT}
Nelson, E. E.~C.; Spence, A.~D.; and Gormley, G.~J. 2023.
\newblock Stepping into the shoes of older people: a scoping review of simulating ageing experiences for healthcare professional students.
\newblock \emph{Age and Ageing}, 52.

\bibitem[{Orr{\`u} et~al.(2023)Orr{\`u}, Piarulli, Conversano, and Gemignani}]{orru2023human}
Orr{\`u}, G.; Piarulli, A.; Conversano, C.; and Gemignani, A. 2023.
\newblock Human-like problem-solving abilities in large language models using ChatGPT.
\newblock \emph{Frontiers in artificial intelligence}, 6: 1199350.

\bibitem[{Park et~al.(2023)Park, O'Brien, Cai, Morris, Liang, and Bernstein}]{park2023generative}
Park, J.~S.; O'Brien, J.; Cai, C.~J.; Morris, M.~R.; Liang, P.; and Bernstein, M.~S. 2023.
\newblock Generative agents: Interactive simulacra of human behavior.
\newblock In \emph{Proceedings of the 36th annual acm symposium on user interface software and technology}, 1--22.

\bibitem[{Park et~al.(2022)Park, Popowski, Cai, Morris, Liang, and Bernstein}]{park2022social}
Park, J.~S.; Popowski, L.; Cai, C.; Morris, M.~R.; Liang, P.; and Bernstein, M.~S. 2022.
\newblock Social Simulacra: Creating Populated Prototypes for Social Computing Systems.
\newblock In \emph{Proceedings of the 35th Annual ACM Symposium on User Interface Software and Technology}, UIST '22. New York, NY, USA: Association for Computing Machinery.
\newblock ISBN 9781450393201.

\bibitem[{Park et~al.(2024)Park, Zou, Shaw, Hill, Cai, Morris, Willer, Liang, and Bernstein}]{park2024generative}
Park, J.~S.; Zou, C.~Q.; Shaw, A.; Hill, B.~M.; Cai, C.; Morris, M.~R.; Willer, R.; Liang, P.; and Bernstein, M.~S. 2024.
\newblock Generative agent simulations of 1,000 people.
\newblock \emph{arXiv preprint arXiv:2411.10109}.

\bibitem[{Pawar et~al.(2021)Pawar, Parkar, Menon, Desai, Namrata, and Dole}]{pawar2021assessment}
Pawar, S.; Parkar, A.; Menon, S.; Desai, N.; Namrata, D.; and Dole, K. 2021.
\newblock Assessment of quality of life of the patients with diabetic retinopathy using National Eye Institute Visual Functioning Questionnaire (VFQ-25).
\newblock \emph{Journal of Healthcare Quality Research}, 36(4): 225--230.

\bibitem[{Project(2024)}]{buffalo2024assistive}
Project, B. A.~A. 2024.
\newblock Multimodal LLM using Federated Visual Instruction Tuning.
\newblock Technical report, University at Buffalo.
\newblock Technical Report.

\bibitem[{Salminen et~al.(2023)Salminen, Jung, Almerekhi, Cambria, and Jansen}]{salminen2023can}
Salminen, J.; Jung, S.-g.; Almerekhi, H.; Cambria, E.; and Jansen, B. 2023.
\newblock How Can Natural Language Processing and Generative AI Address Grand Challenges of Quantitative User Personas?
\newblock In \emph{International Conference on Human-Computer Interaction}, 211--231. Springer.

\bibitem[{SeeingAI(2025)}]{seeingai}
SeeingAI. 2025.
\newblock Seeing AI.

\bibitem[{Shanahan, McDonell, and Reynolds(2023)}]{shanahan2023role}
Shanahan, M.; McDonell, K.; and Reynolds, L. 2023.
\newblock Role play with large language models.
\newblock \emph{Nature}, 623(7987): 493--498.

\bibitem[{Shin et~al.(2024)Shin, Hedderich, Rey, Lucero, and Oulasvirta}]{shin2024understanding}
Shin, J.; Hedderich, M.~A.; Rey, B.~J.; Lucero, A.; and Oulasvirta, A. 2024.
\newblock Understanding human-AI workflows for generating personas.
\newblock In \emph{Proceedings of the 2024 ACM Designing Interactive Systems Conference}, 757--781.

\bibitem[{Stangl et~al.(2021)Stangl, Verma, Fleischmann, Morris, and Gurari}]{stangl2021going}
Stangl, A.; Verma, N.; Fleischmann, K.~R.; Morris, M.~R.; and Gurari, D. 2021.
\newblock Going beyond one-size-fits-all image descriptions to satisfy the information wants of people who are blind or have low vision.
\newblock In \emph{Proceedings of the 23rd international ACM SIGACCESS conference on computers and accessibility}, 1--15.

\bibitem[{Taeb et~al.(2024)Taeb, Swearngin, Schoop, Cheng, Jiang, and Nichols}]{taeb2024axnav}
Taeb, M.; Swearngin, A.; Schoop, E.; Cheng, R.; Jiang, Y.; and Nichols, J. 2024.
\newblock Axnav: Replaying accessibility tests from natural language.
\newblock In \emph{Proceedings of the 2024 CHI Conference on Human Factors in Computing Systems}, 1--16.

\bibitem[{Tripathy and Salini(2025)}]{amsler}
Tripathy, K.; and Salini, B. 2025.
\newblock \emph{Amsler Grid}.
\newblock StatPearls Publishing, Treasure Island (FL).

\bibitem[{Wang et~al.(2023)Wang, Xie, Jiang, Mandlekar, Xiao, Zhu, Fan, and Anandkumar}]{wang2023voyager}
Wang, G.; Xie, Y.; Jiang, Y.; Mandlekar, A.; Xiao, C.; Zhu, Y.; Fan, L.; and Anandkumar, A. 2023.
\newblock Voyager: An open-ended embodied agent with large language models, 2023.
\newblock \emph{URL https://arxiv. org/abs/2305.16291}.

\bibitem[{Wang et~al.(2024)Wang, Ma, Feng, Zhang, Yang, Zhang, Chen, Tang, Chen, Lin et~al.}]{wang2024survey}
Wang, L.; Ma, C.; Feng, X.; Zhang, Z.; Yang, H.; Zhang, J.; Chen, Z.; Tang, J.; Chen, X.; Lin, Y.; et~al. 2024.
\newblock A survey on large language model based autonomous agents.
\newblock \emph{Frontiers of Computer Science}, 18(6): 186345.

\bibitem[{Wang, Zhao, and Kim(2024)}]{wang2024low}
Wang, Y.; Zhao, Y.; and Kim, Y.-S. 2024.
\newblock How Do Low-Vision Individuals Experience Information Visualization?
\newblock In \emph{Proceedings of the 2024 CHI Conference on Human Factors in Computing Systems}, 1--15.

\bibitem[{Xu et~al.(2023)Xu, Wang, Li, Luo, Wang, Liu, and Liu}]{xu2023exploring}
Xu, Y.; Wang, S.; Li, P.; Luo, F.; Wang, X.; Liu, W.; and Liu, Y. 2023.
\newblock Exploring large language models for communication games: An empirical study on werewolf.
\newblock \emph{arXiv preprint arXiv:2309.04658}.

\bibitem[{Zeng et~al.(2020)Zeng, Wang, Chiu, Bhattacharya, and Gurari}]{zeng2020vision}
Zeng, X.; Wang, Y.; Chiu, T.-Y.; Bhattacharya, N.; and Gurari, D. 2020.
\newblock Vision skills needed to answer visual questions.
\newblock \emph{Proceedings of the ACM on Human-Computer Interaction}, 4(CSCW2): 1--31.

\bibitem[{Zhao et~al.(2024)Zhao, Mei, Kumar, and He}]{zhao2024vialm}
Zhao, L.; Mei, S.; Kumar, A.; and He, Y. 2024.
\newblock VIALM: A Survey and Benchmark of Visually Impaired Assistance with Large Models.
\newblock \emph{arXiv preprint arXiv:2402.01735}.

\bibitem[{Zhao et~al.(2018)Zhao, Bennett, Benko, Cutrell, Holz, Morris, and Sinclair}]{zhao2018enabling}
Zhao, Y.; Bennett, C.~L.; Benko, H.; Cutrell, E.; Holz, C.; Morris, M.~R.; and Sinclair, M. 2018.
\newblock Enabling people with visual impairments to navigate virtual reality with a haptic and auditory cane simulation.
\newblock In \emph{Proceedings of the 2018 CHI conference on human factors in computing systems}, 1--14.

\bibitem[{Zhao et~al.(2017)Zhao, Hu, Hashash, and Azenkot}]{zhao2017understanding}
Zhao, Y.; Hu, M.; Hashash, S.; and Azenkot, S. 2017.
\newblock Understanding low vision people's visual perception on commercial augmented reality glasses.
\newblock In \emph{Proceedings of the 2017 CHI conference on human factors in computing systems}, 4170--4181.

\end{thebibliography}

\appendix

\onecolumn

\newpage
\section{Screenshots of Custom Interface for Benchmark of Human Vision Information and Image Perception}\label{apdx:survey_interface}

\begin{figure}[h!]
    \centering
    \includegraphics[width=\linewidth]
    {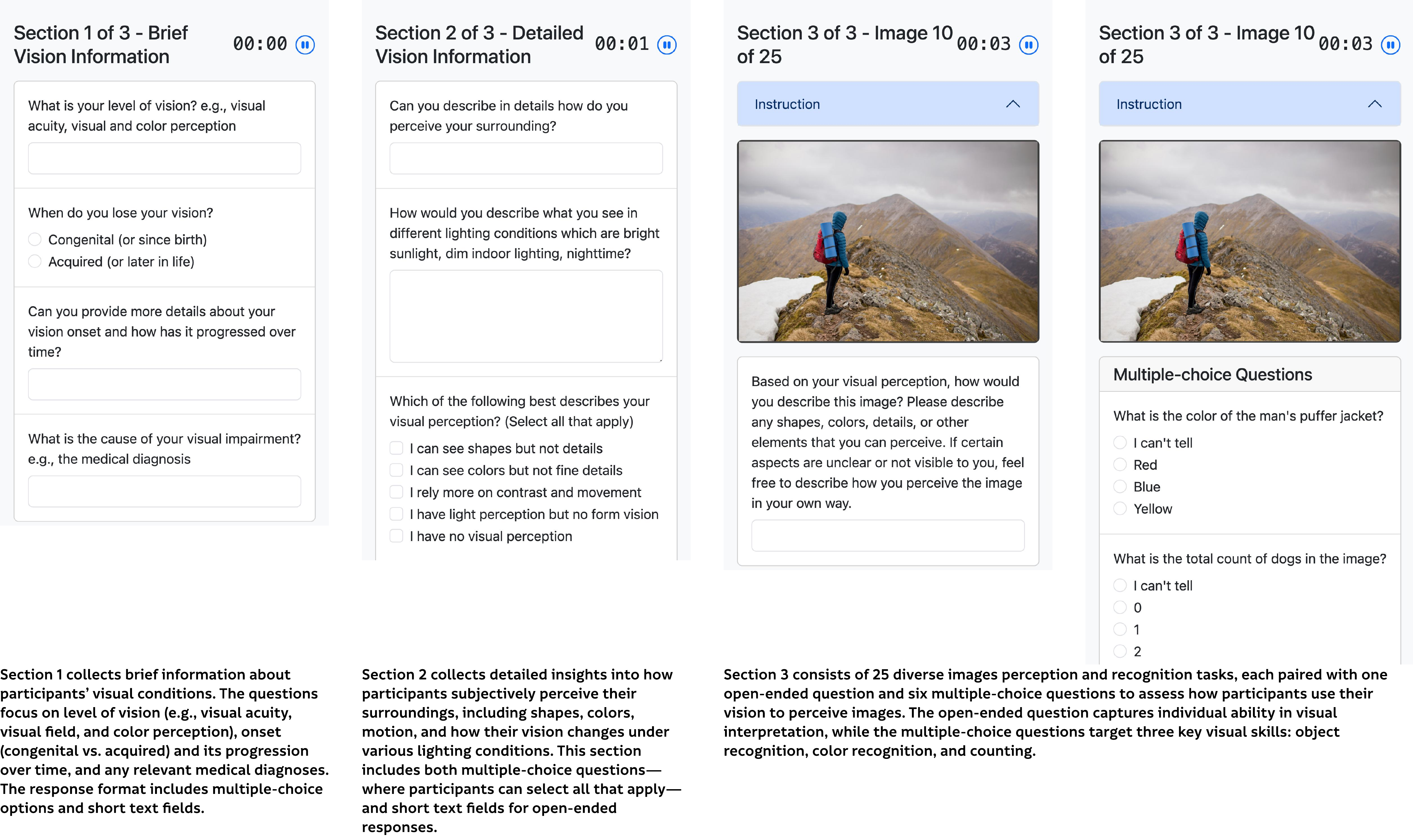}
        \caption{Our custom interface consists of three main sections: (1) Brief Vision Information (Section 1), (2) Detailed Vision Information (Section 2), and (3) Image Perception and Recognition Responses (Section 3). Each section serves a different purpose. Section 1 collects high-level information about the participant’s vision condition. Section 2 gathers more detailed insights into visual perception. Section 3 captures participants’ responses to 25 image perception and recognition tasks.}
    \label{fig:survey_interface}
\end{figure}

\newpage
\section{Collection of Images used for the Benchmark Survey} \label{apdx:image_survey} 
\begin{figure}[h!]
    \centering
    \includegraphics[width=\linewidth]{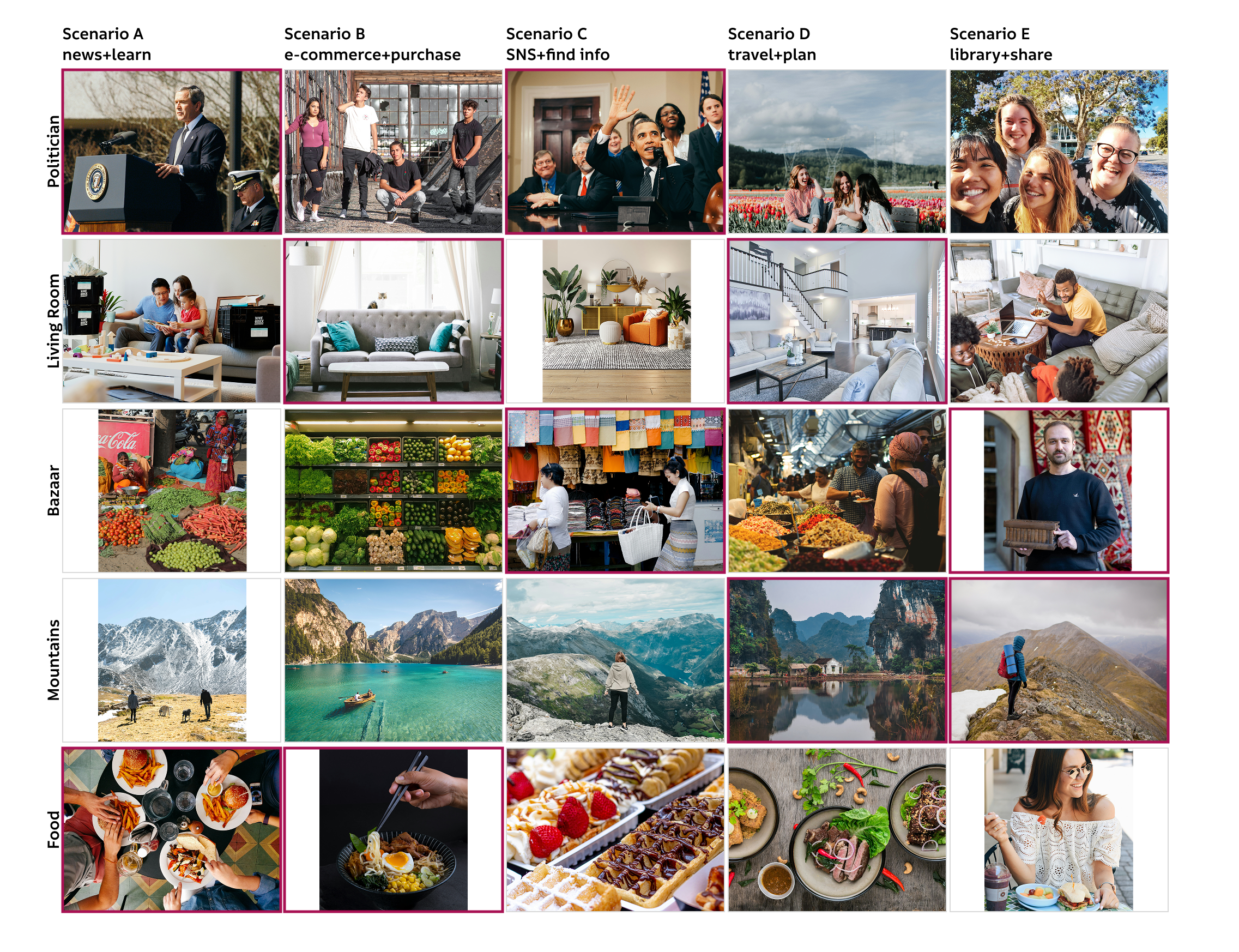}
        \caption{Collections of images that are used for the question in the Image Perception and Recognition section in the survey. In total, there are 25 images that we prepared for the participants, ranging from five scenarios: (1) news + learn, (2) e-commerce + purchase, (3) SNS + find information, (4) travel + planning, and (5) library + sharing information and five types: (i) politician / human, (ii) living room, (iii) bazaar, (iv) mountains, and (v) food. These selections of images are inspired from the types and scenario used in Stangle et. al. In the survey, the participants had to complete the mandatory images before ending the survey that are 10 combinations of images, 2 from each scenario and type (i.e., news-politician, SNS-politician, e-commerce-living room, travel-living room, SNS-bazaar, library-bazaar, travel-mountain, library-mountain, news-food, and e-commerce-food). The mandatory images are highlighted in pink.}
    \label{fig:survey_images}
\end{figure}

Due to the rarity of some combinations (e.g., a politician in an e-commerce setting), and in line with Stangle et. al.'s emphasis on people, their activities, and their surroundings, we substituted these combinations with thematically aligned images.

\newpage
\section{Survey Questions}
\subsection{Demographic Information Section}

\begin{enumerate}
  \item First Name
  \item Last Name
  \item Birth Year
  \item What device are you using currently to complete the survey?
  \begin{enumerate}
    \item Laptop/PC
    \item Mobile Phone
    \item Tablet
  \end{enumerate}
  \item What assistive technologies are you currently using to complete the survey?\\
  \textit{e.g., Screen reader, Magnifier, etc.}
\end{enumerate}

\subsection{Section 1 - Brief Vision Information}

\begin{enumerate}
  \item What is your level of vision?\\
  \textit{e.g., visual acuity, visual and color perception}
  
  \item When did you lose your vision?
  \begin{enumerate}
    \item Congenital (or since birth)
    \item Acquired (or later in life)
  \end{enumerate}
  
  \item Can you provide more details about your vision onset and how it has progressed over time?
  
  \item What is the cause of your visual impairment?\\
  \textit{e.g., the medical diagnosis}
\end{enumerate}

\subsection{Section 2 - Detailed Vision Information}

\begin{enumerate}
  \item Can you describe in detail how you perceive your surroundings?
  \item How would you describe what you see in different lighting conditions (e.g., bright sunlight, dim indoor lighting, nighttime)?
  \item Which of the following best describes your visual perception? (Select all that apply)
  \begin{enumerate}
    \item I can see shapes but not details
    \item I can see colors but not fine details
    \item I rely more on contrast and movement
    \item I have light perception but no form vision
    \item I have no visual perception
  \end{enumerate}
  \item In your experience, do you know if your vision or way of experiencing the world is unique compared to others with the same diagnosis? If so, in what ways?
  \item How much difficulty do you have doing work or hobbies that require you to see well up close, such as cooking, sewing, fixing things around the house, or using hand tools?
  \begin{enumerate}
    \item No difficulty at all
    \item A little difficulty
    \item Moderate difficulty
    \item Extreme difficulty
    \item Stopped doing this because of your eyesight
    \item Stopped doing this for other reasons or not interested in doing this
  \end{enumerate}
  \item How much difficulty do you have finding something on a crowded shelf?
  \begin{enumerate}
    \item No difficulty at all
    \item A little difficulty
    \item Moderate difficulty
    \item Extreme difficulty
    \item Stopped doing this because of your eyesight
    \item Stopped doing this for other reasons or not interested in doing this
  \end{enumerate}
  \item How much difficulty do you have seeing how people react to things you say?
  \begin{enumerate}
    \item No difficulty at all
    \item A little difficulty
    \item Moderate difficulty
    \item Extreme difficulty
    \item Stopped doing this because of your eyesight
    \item Stopped doing this for other reasons or not interested in doing this
  \end{enumerate}
\end{enumerate}

\subsection{Section 3 - Image Perception and Recognition Questions}
\textbf{Note:} The bolded answers are the correct answers.

\begin{enumerate}
    \item \textbf{Bazaar - E-commerce + Purchase}:
        \begin{enumerate}
            \item Q1: What vegetable is in the bottom left corner shelf? \{I can't tell, \textbf{Cabbage}, Cucumber, Carrot\}
            \item Q2: Is there any tomato in this image? \{I can't tell, Yes, \textbf{No}\}
            \item Q3: Is the place indoor or outdoor? \{I can't tell, \textbf{Indoor}, Outdoor\}
            \item Q4: Are there any vegetables that are red in this image? \{I can't tell, \textbf{Yes}, No\}
            \item Q5: What is the total count of people in this image? \{I can't tell, \textbf{0}, 1, 2\}
            \item Q6: What is the total count of boxes that contain banana? \{I can't tell, \textbf{0}, 1, 2\}
        \end{enumerate}

        \item \textbf{Food - E-commerce + Purchase}:
        \begin{enumerate}
        \item Q1: Is there any tomato in the image? \{I can't tell, Yes, \textbf{No}\}
        \item Q2: What cutlery the person is holding in the image? \{I can't tell, Spoon, Fork, \textbf{Chopsticks}\}
        \item Q3: What is the color of the bowl? \{I can't tell, \textbf{Black}, Yellow, Green\}
        \item Q4: What is the color of the meat? \{I can't tell, White, \textbf{Brown}, Red\}
        \item Q5: What is the total count of dogs in the image? \{I can't tell, \textbf{0}, 1, 2\}
        \item Q6: What is the total count of dishes in the image? \{I can't tell, \textbf{1}, 2, 3\}
        \end{enumerate}
        
        \item \textbf{Living Room - E-commerce + Purchase}:
        \begin{enumerate}
        \item Q1: Is there bowl on the table? \{I can't tell, Yes, \textbf{No}\}
        \item Q2: Is there any cat on in the image? \{I can't tell, \textbf{Yes}, No\}
        \item Q3: What is the color of the sofa? \{I can't tell, \textbf{Grey}, Blue, Black\}
        \item Q4: What time of the day is it outside the window? \{I can't tell, \textbf{Day}, Night\}
        \item Q5: What is the total count of cushions are there on the sofa? \{I can't tell, 2, 3, \textbf{5}\}
        \item Q6: What is the total count of lamps are there in the image? \{I can't tell, \textbf{2}, 3, 4\}
        \end{enumerate}

        \item \textbf{Mountains - E-commerce + Purchase}:
        \begin{enumerate}
        \item Q1: Are there any trees in the image? \{I can't tell, \textbf{Yes}, No\}
        \item Q2: Are there mountains in the image? \{I can't tell, \textbf{Yes}, No\}
        \item Q3: What time of the day is the image taken, day or night? \{I can't tell, \textbf{Day}, Night\}
        \item Q4: What is the color of the boat? \{I can't tell, \textbf{Yellow}, Green, Blue\}
        \item Q5: What is the total count of birds in the image? \{I can't tell, \textbf{0}, 1, 2\}
        \item Q6: What is the total count of fishes in the image? \{I can't tell, \textbf{0}, 1, 2\}
        \end{enumerate}
        
        \item \textbf{Politician - E-commerce + Purchase}:
        \begin{enumerate}
        \item Q1: Is there anyone wearing watch? \{I can't tell, \textbf{Yes}, No\}
        \item Q2: What is the man who is looking up holding in his hand? \{I can't tell, \textbf{Black jacket}, Black bag, Black book\}
        \item Q3: What is the weather like? \{I can't tell, Foggy, Rainy, \textbf{Sunny}\}
        \item Q4: What is the color of the girl's shirt? \{I can't tell, Black, \textbf{Pink}, White\}
        \item Q5: What is the total count of people who are standing? \{I can't tell, 2, \textbf{3}, 4\}
        \item Q6: What is the total count of people who are squating down? \{I can't tell, 0, \textbf{1}, 2\}
        \end{enumerate}
        
        \item \textbf{Bazaar - Library + Share}:
        \begin{enumerate}
        \item Q1: What is the material of the item(s) the man is holding? \{I can't tell, Ceramic, \textbf{Wooden}, Plastic\}
        \item Q2: Does the man have any beard? \{I can't tell, \textbf{Yes}, No\}
        \item Q3: What is the color of the man's shirt? \{I can't tell, \textbf{Navy Blue}, Red, Yellow\}
        \item Q4: What is the color of the item the man holding? \{I can't tell, \textbf{Brown}, Red, Blue\}
        \item Q5: What is the total count of the boxes the man is holding on his hand? \{I can't tell, \textbf{1}, 2, 3\}
        \item Q6: What is the total count of cats in the image? \{I can't tell, \textbf{0}, 1, 2\}
        \end{enumerate}
        
        \item \textbf{Food - Library + Share}:
        \begin{enumerate}
        \item Q1: Is the person wearing glasses? \{I can't tell, \textbf{Yes}, No\}
        \item Q2: What is the person holding on her hand? \{I can't tell, Spoon, \textbf{Fork}, Chopsticks\}
        \item Q3: What is the color of her shirt? \{I can't tell, Purple, Green, \textbf{White}\}
        \item Q4: Is the person indoor or outdoor? \{I can't tell, Indoor, \textbf{Outdoor}\}
        \item Q5: What is the total count of the drinks in the image? \{I can't tell, 0, \textbf{1}, 2\}
        \item Q6: What is the total count of cars in the image? \{I can't tell, \textbf{0}, 1, 2\}
        \end{enumerate}
        
        \item \textbf{Living Room - Library + Share}:
        \begin{enumerate}
        \item Q1: What is the cutlery the man is is holding? \{I can't tell, Spoon, \textbf{Fork}, Chopsticks\}
        \item Q2: What is the shape of the coffee table? \{I can't tell, \textbf{Round}, Square, Triangle\}
        \item Q3: What is the color of the kid's pants? \{I can't tell, Yellow, Green, \textbf{Orange}\}
        \item Q4: Is the color of the wall different from the sofa? \{I can't tell, \textbf{Yes}, No\}
        \item Q5: What is the total count of laptops on the table? \{I can't tell, 0, \textbf{1}, 2\}
        \item Q6: What is the total count of phones on the table? \{I can't tell, 0, \textbf{1}, 2\}
        \end{enumerate}
        
        \item \textbf{Mountains - Library + Share}:
        \begin{enumerate}
        \item Q1: What does the person carry on his back? \{I can't tell, Fishing rod, Guitar, \textbf{Backpack}\}
        \item Q2: What is the person doing? \{I can't tell, Sitting down and playing musical instruments overlooking the ocean, \textbf{Standing up in the middle of the mountain and looking at the scenery}, Laying down sideways while reading a book\}
        \item Q3: What is the weather like in the image? \{I can't tell, Snowy, \textbf{Cloudy}, Sunny\}
        \item Q4: What is the color of the man's puffer jacket? \{I can't tell, Red, \textbf{Blue}, Yellow\}
        \item Q5: What is the total count of dogs in the image? \{I can't tell, \textbf{0}, 1, 2\}
        \item Q6: What is the total count of trees in then image? \{I can't tell, \textbf{0}, 2, Many trees\}
        \end{enumerate}
        
        \item \textbf{Politician - Library + Share}:
        \begin{enumerate}
        \item Q1: Is there any car in the image? \{I can't tell, \textbf{Yes}, No\}
        \item Q2: Is there any plane in the image? \{I can't tell, Yes, \textbf{No}\}
        \item Q3: How is the weather shown in the image? \{I can't tell, Snowy, Cloudy, \textbf{Sunny}\}
        \item Q4: What is the hair color of the person on the most left? \{I can't tell, \textbf{Black}, Red, Blonde\}
        \item Q5: What is the total count of people in the image? \{I can't tell, 2, 3, \textbf{4}\}
        \item Q6: What is the total count of people in the image who is wearing glasses? \{I can't tell, 0, \textbf{1}, 2\}
        \end{enumerate}
        
        \item \textbf{Bazaar - News + Learn}:
        \begin{enumerate}
        \item Q1: Does the market sell any meat? \{I can't tell, Yes, \textbf{No}\}
        \item Q2: Is there any car in the image? \{I can't tell, \textbf{Yes}, No\}
        \item Q3: Where is the market located? \{I can't tell, Indoor, \textbf{Outdoor}\}
        \item Q4: What is the weather like in the image? \{I can't tell, \textbf{Sunny}, Rainy, Cloudy\}
        \item Q5: What is the total count of women in this image? \{I can't tell, 0, 1, \textbf{2}\}
        \item Q6: What is the total count of women holding a bottle of water in this image? \{I can't tell, 0, \textbf{1}, 2\}
        \end{enumerate}
        
        \item \textbf{Food - News + Learn}:
        \begin{enumerate}
        \item Q1: Is there anyone have any phone on her/his lap? \{I can't tell, \textbf{Yes}, No\}
        \item Q2: Does everyone have a hamburger on their plate? \{I can't tell, Yes, \textbf{No}\}
        \item Q3: What is the color of the table? \{I can't tell, \textbf{Black}, Yellow, Green\}
        \item Q4: What is the color of the shirt of the person sitting on the left side of the cameraman? \{I can't tell, Pink, Black, \textbf{Grey}\}
        \item Q5: What is the total count of people who are sitting down around the table with visible hands? \{I can't tell, 2, \textbf{3}, 4\}
        \item Q6: What is the total count of cups with beer? \{I can't tell, 0, \textbf{1}, 2\}
        \end{enumerate}
        
        \item \textbf{Living Room - News + Learn}:
        \begin{enumerate}
        \item Q1: Is there any plant in the image? \{I can't tell, \textbf{Yes}, No\}
        \item Q2: What is the item on the table? \{I can't tell, \textbf{Block toys}, Books, Flowers\}
        \item Q3: What is the color of the big box on the sofa? \{I can't tell, Red, \textbf{Black}, White\}
        \item Q4: What is the color of the kid's shirt? \{I can't tell, \textbf{Red}, Black, White\}
        \item Q5: What is the total count of people in the image? \{I can't tell, 2, \textbf{3}, 4\}
        \item Q6: What is the total count of cups the man is holding? \{I can't tell, 0, \textbf{1}, 2\}
        \end{enumerate}
        
        \item \textbf{Mountains - News + Learn}:
        \begin{enumerate}
        \item Q1: Is there any snow on the ground? \{I can't tell, \textbf{Yes}, No\}
        \item Q2: What are the people doing in the image? \{I can't tell, \textbf{Walking towards the mountain.}, Laying down on the ground., Chatting while sitting down.\}
        \item Q3: What is the weather like in the image? \{I can't tell, \textbf{Sunny}, Rainy, Cloudy\}
        \item Q4: What is the color of the dog on the right? \{I can't tell, \textbf{Black}, Brown, White\}
        \item Q5: What is the total count of the dogs in the image? \{I can't tell, 0, 1, \textbf{2}\}
        \item Q6: What is the total count of people in the image? \{I can't tell, \textbf{2}, 3, 4\}
        \end{enumerate}
        
        \item \textbf{Politician - News + Learn}:
        \begin{enumerate}
        \item Q1: Is this indoor or outdoor? \{I can't tell, Indoor, \textbf{Outdoor}\}
        \item Q2: What is the weather like? \{I can't tell, \textbf{Sunny}, Rainy, Stormy\}
        \item Q3: Are there any tree in the image? \{I can't tell, \textbf{Yes}, No\}
        \item Q4: Is the man standing on the podium wearing glasses? \{I can't tell, Yes, \textbf{No}\}
        \item Q5: What is the total count of people in the image? \{I can't tell, 1, 2, \textbf{3}\}
        \item Q6: What is the total count of the microphones in the image? \{I can't tell, \textbf{2}, 3, 4\}
        \end{enumerate}
        
        \item \textbf{Bazaar - SNS + Find Info}:
        \begin{enumerate}
        \item Q1: Is the person on the most left carrying a bag? \{I can't tell, \textbf{Yes}, No\}
        \item Q2: Which direction the people is walking towards in the image? \{I can't tell, Towards the camera, \textbf{Left}, Right\}
        \item Q3: What is the color of the shirt of the person on the most right? \{I can't tell, \textbf{White}, Yellow, Green\}
        \item Q4: What is the color of the bag which the person on the most right holding? \{I can't tell, \textbf{White}, Black, Orange\}
        \item Q5: What is the total count of individuals with their hair tied? \{I can't tell, 0, \textbf{1}, 2\}
        \item Q6: What is the total count of people in the image? \{I can't tell, \textbf{2}, 3, 4\}
        \end{enumerate}
        
        \item \textbf{Food - SNS + Find Info}:
        \begin{enumerate}
        \item Q1: What kind of food is shown in the image? \{I can't tell, Savory, \textbf{Dessert}\}
        \item Q2: Is there any cutlery in the image? \{I can't tell, Yes, \textbf{No}\}
        \item Q3: What is the red fruit topping? \{I can't tell, Cherry, \textbf{Strawberry}, Watermelon\}
        \item Q4: What is the color of the container of the food? \{I can't tell, Red, Black, \textbf{White}\}
        \item Q5: What is the total count of people in the image? \{I can't tell, \textbf{0}, 1, 2\}
        \item Q6: What is the total count of the apples in the image? \{I can't tell, \textbf{0}, 1, 2\}
        \end{enumerate}
        
        \item \textbf{Living Room - SNS + Find Info}:
        \begin{enumerate}
        \item Q1: Are there any lamps in the image? \{I can't tell, \textbf{Yes}, No\}
        \item Q2: What is the pattern of the flooring? \{I can't tell, \textbf{Wooden}, Stone, Concrete\}
        \item Q3: What is the color of the armchair? \{I can't tell, \textbf{Orange}, Black, White\}
        \item Q4: What is the color of the wall? \{I can't tell, \textbf{Beige}, Orange, Green\}
        \item Q5: What is the total count of the armchairs in the image? \{I can't tell, 0, \textbf{1}, 2\}
        \item Q6: What is the total count of mirrors in the image? \{I can't tell, 0, \textbf{1}, 2\}
        \end{enumerate}
        
        \item \textbf{Mountains- SNS + Find Info}:
        \begin{enumerate}
        \item Q1: What is the background scenery of the image? \{I can't tell, Flower Park, Waterfall, \textbf{Mountain}\}
        \item Q2: Where is the person facing? \{I can't tell, \textbf{Looking at the background}, Looking at the camera, Looking at the side of the image\}
        \item Q3: What is the weather like? \{I can't tell, Foggy, Rainy, \textbf{Cloudy}\}
        \item Q4: What color is the girl's hoodies? \{I can't tell, \textbf{Grey}, Black, Blue\}
        \item Q5: What is the total count of dogs in the image? \{I can't tell, \textbf{0}, 1, 2\}
        \item Q6: What is the total count of birds in the image? \{I can't tell, \textbf{0}, 1, 2\}
        \end{enumerate}
        
        \item \textbf{Politician - SNS + Find Info}:
        \begin{enumerate}
        \item Q1: Is there anyone holding a phone? \{I can't tell, \textbf{Yes}, No\}
        \item Q2: What is the mood of the people? \{I can't tell, \textbf{Happy}, Sad\}
        \item Q3: What is the color of the phone? \{I can't tell, \textbf{Black}, Brown, Green\}
        \item Q4: What color is the suit of the man holding the phone? \{I can't tell, Blue, \textbf{Black}, Red\}
        \item Q5: What is the total count of individuals who are raising their hands? \{I can't tell, 0, \textbf{1}, 2\}
        \item Q6: What is the total count of people who are sitting down? \{I can't tell, 2, \textbf{3}, 4\}
        \end{enumerate}

        \item \textbf{Bazaar - Travel + Plan}:
        \begin{enumerate}
        \item Q1: What kind of stuff is sold in the place? \{I can't tell, Clothes, \textbf{Food}, Vases\}
        \item Q2: What is the pattern of shirt of the man with glasses and holding a plate on his hand? \{I can't tell, Floral, Stripes, \textbf{Checkered}\}
        \item Q3: Is there anyone in the image wearing a pink shirt? \{I can't tell, \textbf{Yes}, No\}
        \item Q4: What is the color of the backpack the woman with headgear? \{I can't tell, Red, \textbf{Black}, White\}
        \item Q5: What is the total count of people holding a plate in this image? \{I can't tell, 0, \textbf{1}, 2\}
        \item Q6: What is the total count of cars in the image? \{I can't tell, \textbf{0}, 1, 2\}
        \end{enumerate}
        
        \item \textbf{Food - Travel + Plan}:
        \begin{enumerate}
        \item Q1: What is the food placed inside the center bowl? \{I can't tell, Rice, \textbf{Meat}, Noodle\}
        \item Q2: Is there any nut in the image? \{I can't tell, \textbf{Yes}, No\}
        \item Q3: What is the color of the chili in the image? \{I can't tell, \textbf{Red}, Green, Yellow\}
        \item Q4: What is the color of the table? \{I can't tell, \textbf{Brown}, Yellow, Red\}
        \item Q5: What is the total count of bowls in the image? \{I can't tell, 1, \textbf{4}, 5\}
        \item Q6: What is the total count of chilies on the big bowl in the middle? \{I can't tell, \textbf{1}, 4, 5\}
        \end{enumerate}
        
        \item \textbf{Living Room - Travel + Plan}:
        \begin{enumerate}
        \item Q1: Is there any painting on the wall in this image? \{I can't tell, \textbf{Yes}, No\}
        \item Q2: Is there any book on the table? \{I can't tell, Yes, \textbf{No}\}
        \item Q3: What is the color of the sofa? \{I can't tell, \textbf{White}, Yellow, Blue\}
        \item Q4: Is the light on? \{I can't tell, \textbf{Yes}, No\}
        \item Q5: What is the total count of people in the image? \{I can't tell, \textbf{0}, 3, 5\}
        \item Q6: What is the total count of cats in the image? \{I can't tell, \textbf{0}, 1, 2\}
        \end{enumerate}
        
        \item \textbf{Mountains - Travel + Plan}:
        \begin{enumerate}
        \item Q1: Is there any duck in the lake? \{I can't tell, Yes, \textbf{No}\}
        \item Q2: Is there any lake in the image? \{I can't tell, \textbf{Yes}, No\}
        \item Q3: What is the weather like? \{I can't tell, Snowy, \textbf{Cloudy}, Raining\}
        \item Q4: What is the color of the roof of the most right house? \{I can't tell, Green, \textbf{Dark Brown}, Yellow\}
        \item Q5: What is the total count of houses with an aged or off-white appearance in the image? \{I can't tell, 0, \textbf{1}, 2\}
        \item Q6: What is the total count of people in the image? \{I can't tell, \textbf{0}, 1, 2\}
        \end{enumerate}
        
        \item \textbf{Politician - Travel + Plan}:
        \begin{enumerate}
        \item Q1: Is there any bird in the image? \{I can't tell, Yes, \textbf{No}\}
        \item Q2: What are the facial expressions of the people sitting on the bench? \{I can't tell, \textbf{Happy and engaged in conversation}, Serious and deep in thought, Bored and uninterested\}
        \item Q3: Is the image taken indoor or outdoor? \{I can't tell, Indoor, \textbf{Outdoor}\}
        \item Q4: What is the hair color of the girl sitting at the most right of the bench? \{I can't tell, Blond, \textbf{Black}, Red\}
        \item Q5: What is the total count of people who are sitting down on the bench and having conversation? \{I can't tell, 2, \textbf{3}, 4\}
        \item Q6: What is the total count of balloons in the image? \{I can't tell, \textbf{0}, 3, 5\}
        \end{enumerate}
\end{enumerate}
% \label{apdx:survey_questions}

\newpage
\section{Prompt Template}
The figure illustrates the prompt template used in our analysis. The template is organized into three parts: (1) a system prompt that describes the vision context from participant information and the role the agent should adopt; (2) a user prompt that states the task instruction and may or may not include example responses for image perception and recognition tasks; and (3) a closing instruction that specifies the format and content expected from the VLMs. To assemble a complete prompt, the user selects one option from the dashed rectangles contained within each corresponding solid rectangle in the diagram.

\begin{figure*}[h!]
    \centering
    \includegraphics[width=11.5cm]{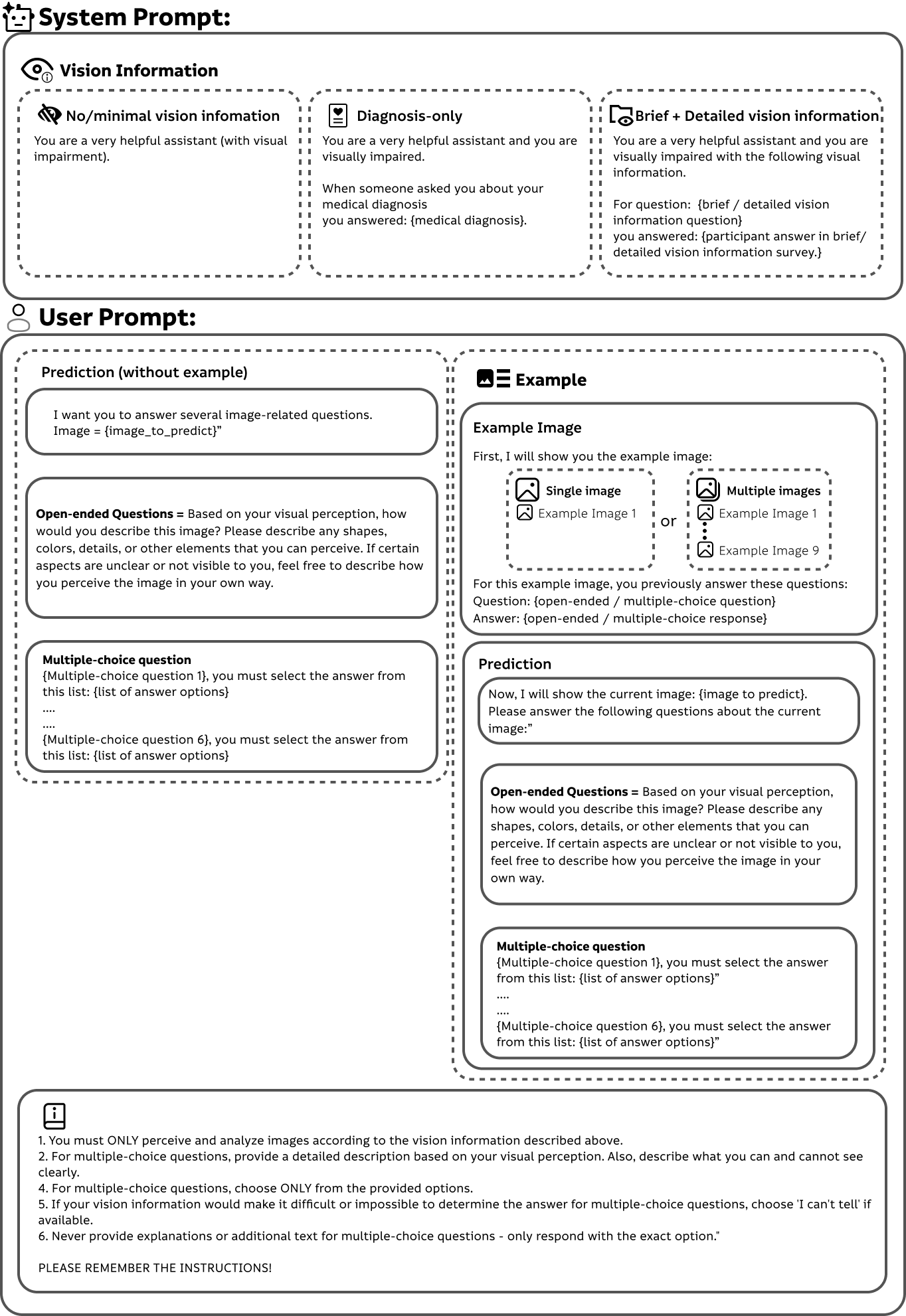}
    \caption{The figure shows the prompt template used in our analysis. Overall, each prompt consists of two main components: a system prompt and a user prompt. The system prompt can include vision information at one of three levels of detail: (1) No or minimal vision information, which serves as a baseline with no vision-related content; (2) Diagnosis-only, which includes a brief diagnostic statement; and (3) Brief + Detailed vision information, which provides a full vision profile derived from the participant’s response in the Detailed Vision Information section. The user prompt provides prediction instructions (with or without example inclusion) and prediction format. In the figure, dashed rectangles represent optional components within the solid rectangles components.}
    \label{fig:prompt_template}
\end{figure*}

\onecolumn
\section{Prompt Combinations used for Agent Evaluation}
As discussed in Section 4.1, we design 16 prompts for our performance analysis. These prompts are created by combining different combinations of vision information, example, and type of responses. Below, we will present all the prompt combinations used our analysis, following the template shown:

\begin{verbatim}
    Prompt X = {Vision Information}{Example}{Type of Responses}
\end{verbatim} 

The field \texttt{Vision Information} may include one or more items from \texttt{{diagnosis, brief, detailed}}. The \texttt{Example} field may contain one or more items from \texttt{{unrelated, type, scenario, multiple}}. The \texttt{Type of Responses} field may include either one or both of \texttt{{open-ended, multiple-choice}}. An empty set of curly brackets indicates that the prompt does not include any item for that specific field.

\subsection{Prompt Combinations}
\begin{quote}
\begin{verbatim}
Prompt 1 = {diagnosis}{}{}
Prompt 2 = {brief}{}{}
Prompt 3 = {brief, detailed}{}{}
Prompt 4 = {}{}{open-ended, multiple-choice}
Prompt 5 = {brief, detailed}{unrelated}{open-ended}
Prompt 6 = {brief, detailed}{unrelated}{multiple-choice}
Prompt 7 = {brief, detailed}{unrelated}{open-ended, multiple-choice}
Prompt 8 = {brief, detailed}{type}{open-ended}
Prompt 9 = {brief, detailed}{type}{multiple-choice}
Prompt 10 = {brief, detailed}{type}{open-ended, multiple-choice}
Prompt 11 = {brief, detailed}{scenario}{open-ended}
Prompt 12 = {brief, detailed}{scenario}{multiple-choice}
Prompt 13 = {brief, detailed}{scenario}{open-ended, multiple-choice}
Prompt 14 = {brief, detailed}{multiple}{open-ended}
Prompt 15 = {brief, detailed}{multiple}{multiple-choice}
Prompt 16 = {brief, detailed}{multiple}{open-ended, multiple-choice}
\end{verbatim}
\end{quote}

\end{document}